\documentclass[
twocolumn,
hf,
]{ceurart}

\usepackage{listings}
\usepackage{lettrine}
\usepackage{subcaption}
\usepackage{bm}
\begin{document}



\title{Enhancing robot reliability for health-care facilities by means of Human-Aware Navigation Planning}


\author[1]{Olga E. Sorokoletova}[%
email=olgasorokoletova02@gmail.com,
]
\cormark[1]
\address[1]{Sapienza University of Rome,
  5 Piazzale Aldo Moro, Rome RM, 00185, Italy}

\author[1]{Luca Iocchi}[%
]


\begin{abstract}
With the aim of enabling robots to cooperate with humans, carry out human-like tasks, or navigate among humans, we need to ensure that they are equipped with the ability to comprehend human behaviors and use the extracted knowledge for intelligent decision-making. This ability is particularly important in the safety-critical and human-centred environment of health-care institutions. In the field of robotic navigation, the most cutting-edge approaches to enhancing robot reliability in the application domain of healthcare facilities and in general pertain to augmenting navigation systems with human-aware properties. To implement this in our work, the Co-operative Human-Aware Navigation planner has been integrated into the ROS-based differential-drive robot MARRtina and exhaustively challenged within various simulated contexts and scenarios (mainly modelling the situations relevant in the medical domain) to draw attention to the integrated system's benefits and identify its drawbacks or instances of poor performance while exploring the scope of system capabilities and creating a full characterization of its applicability. The simulation results are then presented to medical experts, and the enhanced robot acceptability within the domain is validated with them as the robot is further planned for deployment.
\end{abstract}


\maketitle

\section{Introduction}
\lettrine[lines=3, findent=2pt, nindent=-3pt]{T}{}aking into account the specificity of the application domain is one of the crucial considerations while developing a robotic system, and the chosen domain of a health-care facilities is represented by the quick-paced and safety-critical environment, which is frequently overcrowded, chaotic, and understaffed. The robotics community has been researching the use of robots in hospital settings to reduce the workload of caregivers by utilizing robots for low-value duties such as medical supply delivery or patient assistance at the bedside when clinicians are not available.

Nowadays, the number of autonomous mobile robots serving in hospitals is rather low. This reveals a major gap in terms of existing approaches and is explained by the fact that the higher the desired level of autonomy, the more challenging it is to provide it while also satisfying safety criteria of primary importance for the considered application domain. Thus, the clinical environment is a unique \textbf{safety-critical environment} that poses specific navigation challenges for robots \cite{ref:situating}, \cite{ref:safedqn} and gives rise to particular scenarios in which the navigational conflicts between humans and robots must be resolved.

For example, a robot in an Intensive Care Unit (ICU) is likely to encounter the situation when a group of Health-Care Workers (HCWs), some of whom may have been involved in other tasks and whose predictions of behaviors may have already been derived by the robot's planner, suddenly rush to the bed of a certain high-acuity patient to perform a life-saving treatment. In this scenario, to address the challenge of performing the situation assessment and responding to the changes in dynamics of the environment due to changing human plans, robots must be more than purely reactive; they must be \textbf{proactive}, \textbf{adaptive}, \textbf{anticipative}, and \textbf{capable of intelligent decision-making}.

Besides that, robot \textbf{adaptability} is required, which is defined as the ability of a robot to adapt to a new environment with different characteristics. To illustrate, let us assume that in a particular hospital, the ICU is a room on the Emergency Department (ED) floor. The ICU resembles a crowded open-space kind of environment (beds are normally installed along the walls, and often there is a wall and/or nursing station in the center of the room); meanwhile, the corridors of the ED could be not so crowded but cluttered with trolleys and appliances. Therefore, while exiting the ICU to navigate in the ED hallways and vice versa, the robot must adapt to changes and exhibit different types of behavior.

\begin{figure}
  \centering
  \includegraphics[width=.3\linewidth]{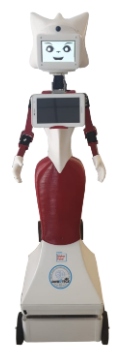}
  \caption{MARRtina bedside robot. Photograph from the gallery of the robot web page. (\url{https://www.marrtino.org/robots}).}
  \label{fig:marrtina}
\end{figure}

In our work, we address the described challenges and also the challenge of increasing the robot's \textbf{acceptability} and mitigating inconsistencies between the robot's actual and expected behavior by means of Human-Aware Navigation Planning. Specifically, we integrate the \textbf{Co-operative Human-Aware Navigation (CoHAN) planner} designed by Singamaneni, Favier, and Alami \cite{ref:cohan} into a given robot software framework. The CoHAN planner enables a robot to navigate in diverse contexts while being aware of humans in the environment. The software belongs to a ROS-based differential drive robot with functionalities equivalent to those of a Turtlebot system. By now, experimentation has been carried out in a simulation, but the social version of the robot in \autoref{fig:marrtina}, MARRtina, is intended for deployment in a real hospital\footnote{Sant'Andrea Hospital, Roma RM, Italy} to serve as a bedside robot.

The integrated system becomes capable of performing \textbf{Social Navigation} as opposed to treating people simply as obstacles. The studies within the Social (or Human-Aware) Navigation field are centered on learning about various \textbf{human motion patterns} and relevant \textbf{social aspects} and developing navigation approaches that account for them. The associated research community has explored a wide range of interesting subject matters so far and developed numerous Human-Aware Navigation planners; however, these planners are mostly environment- and scenario-specific and hence do not meet the adaptability requirement indicated above. Therefore, the motivation behind choosing the CoHAN system for integration is that it is \textbf{flexible}, highly \textbf{tunable}, and easily \textbf{adjustable} to the various contexts, making it capable of handling complex indoor and crowded scenarios (ICU, ED).

The main contribution of this work is that, in addition to being integrated, a new planning system, enhanced with human awareness, has been thoroughly challenged, tested, and evaluated in multiple simulated human-robot scenarios and environments: ICU crowded free space, ICU with a wall in the middle, heart attack emergency in ICU, narrow hallway, free/cluttered wide ED corridor, narrow/wide door crossing scenario, patient bed approach scenario, etc. Qualitative and quantitative analysis is presented, uncovering the subtle nuances of the system's functioning and ultimately creating a \textbf{comprehensive characterization of the properties and limitations of the adopted planning approach}. As such, on a global level, we hope to contribute to better patient outcomes and lessen the workload of HCWs by making robots more \textbf{reliable}. One of the necessary steps towards this goal is to ensure that potential users accept the developed robotic technology. Therefore, after gathering feedback on the system's performance from the HCWs at the Sant'Andrea Hospital, a brief \textbf{statistical analysis of the robot's acceptability} by humans was carried out.

\bigskip
The rest of the paper is structured as follows. The background material and related works are described in \hyperref[sec:2]{Section~\ref*{sec:2}}. The CoHAN Planner’s architecture, including an overview of its features, components, and the scheme of communication between different modules and processes, is provided in \hyperref[sec:3]{Section~\ref*{sec:3}}. Following this, \hyperref[sec:4]{Section~\ref*{sec:4}} deals with the details of the integration of a CoHAN system into MARRtina software. The primary matter of the work is found in \hyperref[sec:5]{Section~\ref*{sec:5}}. It covers the experimental set-up and analysis of the results in various simulated human-robot scenarios. Finally, the work is summarized in \hyperref[sec:6]{Section~\ref*{sec:6}}, which also comments on potential future research to improve the system.

\section{Background}
\label{sec:2}
This section briefly explains the fundamentals and some related works of the Robotic Operation System (ROS) 2D Navigation Stack\footnote{\url{http://wiki.ros.org/navigation}}, in general, and a package with a Timed Elastic Band (TEB) local trajectory planner\footnote{\url{http://wiki.ros.org/teb\_local\_planner}}, in particular, because this is exactly the component that has been made "human-aware" in the architecture of the CoHAN system. Then, the Human-Aware Navigation Planning problem is introduced.

\subsection{ROS Navigation Stack}
\label{subsec:stack}
Both MARRtino robot software and the integrated Co-operative Human-Aware Navigation (CoHAN) planner are ROS-based. Citing its web page, ROS \cite{ref:ros} is an open-source robotic middleware toolset that provides a structured communications layer above the host operating systems and is considered by many as the most general-purpose and commonly-used frameworks for developing robotic applications.

The atomic units for organizing software in ROS are \textbf{packages}, and the collections of packages are a \textbf{stack}. Particularly, to navigate with a ROS robot, the ROS Navigation Stack is used. The ROS navigation, or ROS Navigation Stack, is meant for 2D maps, and it takes as input information about \textbf{odometry}, \textbf{sensor measurements}, and a \textbf{goal pose} to provide as output safe \textbf{velocity commands} that are sent to a mobile base.

Functionally, the Navigation Stack packages contain a set of ROS nodes and navigation-related algorithms that are implemented to move mobile robots autonomously from one point to another while avoiding obstacles. The basic structure of a Navigation Stack is represented by the three main blocks or modules:
\begin{itemize}
\item \textbf{Mapping} $-$ to create a map of the environment;
\item \textbf{Localization} $-$ to localize a robot in the created map;
\item \textbf{Path Planning} $-$ to plan the robot path (Global Path Planning) and execute the planned path while avoiding obstacles (Local Path Planning).
\end{itemize}

\subsection{Timed Elastic Band}
\label{subsec:teb}
The main accent in a newly integrated CoHAN system is the improvement of a particular component of the described ROS Navigation Stack: the local planner. Precisely, the Timed Elastic Band (TEB) is the \textbf{optimal local trajectory planner}. This planner's package is called \verb|teb_elastic_band|and is implemented in ROS as a plugin to the default local planner of the 2D Navigation Stack. The underlying planning approach was introduced in \cite{ref:tebart}, and its core idea is to optimize locally at runtime the initial trajectory generated by a global planner with respect to:
\begin{itemize}
\item \textbf{Trajectory execution time} (time-optimal objective);
\item \textbf{Separation from the obstacles};
\item \textbf{Kinodynamic constraints} (maximum velocities and accelerations).
\end{itemize}

The Elastic Band is a sequence of $n$ intermediate robot configurations $\bm{x_i} = (x_i, y_i, \beta_i)^T \in R^2 \times S^1$, where $(x_i, y_i)$ is a position and $\beta_i$ $-$ orientation of the robot in a map reference frame:

\begin{equation}
    Q = \{\bm{x_i}\}_{i=0...n} \quad n \in \mathbb{N}
\end{equation}
\smallskip

The \textbf{Timed Elastic Band (TEB)} appears as an augmentation by the time intervals between two consecutive configurations: $n-1$ time differences $\Delta T_i$, each of which is the time needed to transit from one configuration to another, i.e., as a tuple of sequences:

\begin{equation}
\label{eq:tuple}
    B := (Q, \tau) \quad \tau = \{\Delta T_i\}_{i=0...n-1}
\end{equation}
\smallskip

Then, the optimization problem can be formulated as a \textbf{real-time weighted multi-objective optimization of TEB} defined in \autoref{eq:tuple} in terms of both configurations and time intervals. Let us denote $f(B)$ as an objective function and $B^*$ as an optimal value (problem solution), then:
\begin{align}
    f(B) = \sum_k \gamma_kf_k(B) \\
    B^* = \underset{B}{\arg\min} {f(B)}
\end{align}
where $\gamma_k$ are the weights and $f_k$ $-$ multiple objectives that may capture the diverse aspects.

\subsection{Human-Aware Navigation}
\label{subsec:hanav}
The TEB planner was further expanded with human-aware characteristics to get \textbf{Human-Aware Timed Elastic Band (HATEB)} \cite{ref:hateb} by incorporating human motion predictions and estimations and simultaneous planning for humans and the robot. By "human-aware characteristics", we mean that the robot's paths generated by the planner must not only be safe, legible, and optimal in terms of time and robot resource consumption, but also acceptable and look natural to humans. And Social (or Human-Aware) Navigation is a whole new branch of robotic research that emerged from the synthesis of Human-Robot Interaction (HRI) and Motion Planning to focus on learning how to build such paths.

A survey in \cite{ref:hanav} suggests that the Human-Aware Navigation challenge can be defined as a challenge of navigating while accounting for the \textbf{constraints related to social aspects and rules} and offers one of the possible approaches to the sub-categorization of the characteristics that a robot must exhibit in order to be considered navigating in a human-aware manner: compliance with human comfort, naturalness of motion, and sociability.

\begin{figure}
    \centering
    \includegraphics[width=0.8\linewidth]{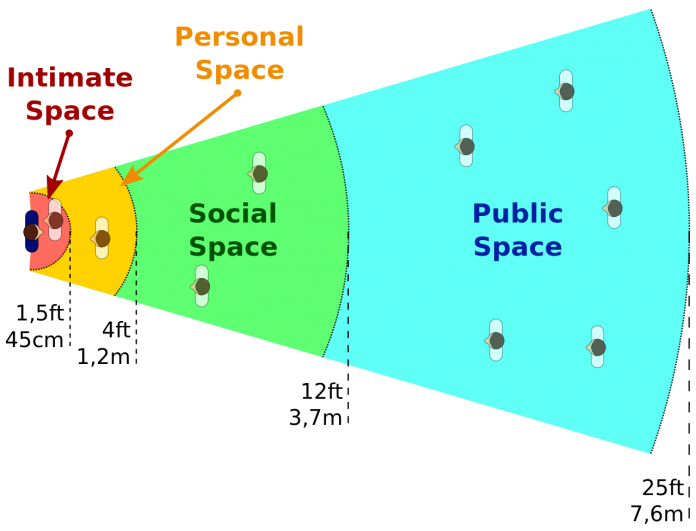}
    \caption{\centering \textbf{PROXIMITY}: \textbf{Intimate distance} – indicates a closer relationship; \textbf{Personal distance} – occurs between family members or close friends; \textbf{Social distance} – used with individuals who are acquaintances; \textbf{Public distance} – used in public speaking situations. Image credit: Jean-Louis Grall, via \href{https://commons.wikimedia.org/wiki/File:Personal_Spaces_in_Proxemics.svg}{Wikimedia
    Commons}.}
    \label{fig:proxemics}
\end{figure}

\textbf{Human Comfort} appears as the absence of annoyance and stress for humans during human-robot interaction sessions or in shared spaces. In a sense, it is linked to the concept of safety and heavily relies on the study of \textbf{proxemics}, the study of how people perceive the proximity of others, illustrated in \autoref{fig:proxemics}. It is a branch of knowledge that deals with the amount of space that people feel it is necessary to set between themselves and others to feel comfortable.

\textbf{Motion Naturalness} is the ability of the robot to mimic the nature of human motion and elaborate human-alike navigation behavior by capturing and recreating human low-level motion patterns. The main struggle in this research area resides in the fact that robots have different abilities compared to humans, and therefore not all patterns are meant to be transferred.

Finally, \textbf{Sociability} is concerned with high-level decision-making and attempts to comply with social and cultural norms by either explicit modeling of known social protocols or learning them from humans as spacial effect. The latter is especially advantageous because learning from humans allows their behavior to be not only predicted but also affected and exploited for spacial conflict resolution.

\section{Co-operative Human-Aware Navigation Planner Architecture}
\label{sec:3}
\begin{figure}
    \centering
    \includegraphics[width=\linewidth]{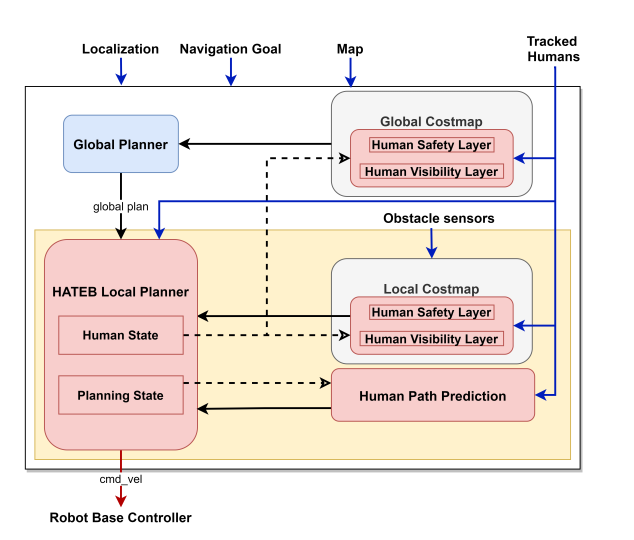}
    \caption{\centering CoHAN planner software architecture. Image credit: Singamaneni et al, \cite{ref:cohan}.}
    \label{fig:architecture}
\end{figure}

As introduced previously, the Co-operative Human-Aware Navigation (CoHAN) planner is a chosen approach to implementing Social Navigation in our system. Its overall architecture is summarized in the block scheme in \autoref{fig:architecture} and described in the current section.

The red building blocks in the block scheme are the CoHAN-specific components that were added by Singamaneni et al. \cite{ref:cohan} to the standard ROS Navigation Stack to enrich it with human-aware properties. The HATEB Local Planner is a core component of the system, which is in essence a human-aware extension of the Timed Elastic Band (TEB) local planner, explained in \hyperref[subsec:teb]{Section~\ref*{subsec:teb}}. It includes two state variables: the Human State and the Planning State. 

Human State is a part of the Human Tracking mechanism that controls the inclusion of the Human Safety and Human Visibility costmap layers, where the former adds cost for approaching humans too close and the latter penalizes surprise appearances from behind and makes the robot enter the human’s field of view from a larger distance.

Planning State is a criterion for the selection of a Human Path Prediction service, whereas a Human Path Prediction module is responsible for sending a request to a certain service in order to predict or suggest possible paths for the tracked dynamic humans.

To better understand the interconnections between the different components of architecture, let us describe the Human Tracking approach, the Human Path Prediction module, and the transition scheme between Planning Modes in more detail.

\subsection{Human Tracking}
\label{subsec:tracking}
The Human Tracking is provided by an external to the Navigation Stack module (top-right text input in the \autoref{fig:architecture}). Although all humans in the environment are tracked, only those within a Visible Region are considered visible to the robot. Furthermore, by introducing the Planning Radius (a tunable parameter), the Visible Region is further narrowed, and only humans within the Planning Radius are considered for planning.

\begin{figure}[h]
\centering
\begin{minipage}[h]{0.49\linewidth}
\center{\includegraphics[width=.7\linewidth]{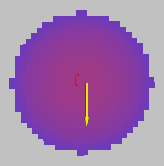} \\ Human Safety.}
\end{minipage}
\begin{minipage}[h]{0.49\linewidth}
\center{\includegraphics[width=.7\linewidth]{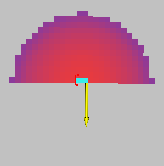} \\ Human Visibility.}
\end{minipage}
\caption{\centering Costmap layers around the static humans, displayed in RViz. The yellow arrow denotes a human pose. The Human Safety layer (blue) is a 2D Gaussian around the human, and the Human Visibility layer (red) is a 2D half Gaussian on the backside of the human. Both of these layers have a cutoff radius beyond which the cost is zero.}
\label{fig:costmap}
\end{figure}

The humans within the Planning Radius are called \textbf{observable}. Each observable human is classified as "static" or "dynamic," and the classification result is recorded in the Human State. Following that, the costmaps' plugin examines the Human State and adds the Human Safety and Human Visibility costmap layers around the \textbf{static} humans as it is shown in \autoref{fig:costmap}.

When it comes to \textbf{dynamic} humans, the system detects the two nearest of them, attaches \textbf{elastic bands}, predicts their paths, and plans their possible trajectories until they move behind the robot or out of the Planning Radius.

\subsection{Human Path Prediction}
\label{subsec:prediction}
The Human Path Prediction module is responsible for the generation of global plans for the two nearest observable dynamic humans. Once the global plans are generated, they are handed over to the HATEB Local Planner to build the local plans.

There are four different prediction services, and which one of them is currently in charge depends on the system configuration and active Planning Mode. The services are called \verb|PredictBehind|, \verb|PredictGoal|, \verb|PredictVelObs| and \verb|PredictExternal|. They are described in detail in \cite{ref:cohan}. However, our experimentation allowed us to discover the nuances of their functioning, as reported in the following paragraph.

\verb|PredictBehind|, \verb|PredictGoal|, and \verb|PredictExternal| services are called when the robot is in a \texttt{DualBand} mode. \verb|PredictBehind| and \verb|PredictGoal| are alternatives to each other, and \verb|PredictExternal| can be activated in parallel with any of them. Activation is done manually through the configuration file. Controversially, \verb|PredictVelObs| is called automatically when the robot switches to \texttt{VelObs} mode. Note that \verb|PredictBehind|, \verb|PredictGoal|, and \verb|PredictExternal| can all be disabled at the same time, and then \verb|PredictVelObs| is used for \verb|DualBand| as well.

\subsection{Planning Modes}
\label{subsec:modes}
The CoHAN system is designed to handle a variety of contexts. For that, a \textbf{multi-modality} is provided. The concept of multi-modality, or modality shifting, consists of a mechanism of switching between different Planning Modes depending on the current context. This concept can be related to the ideas of Qian et al. \cite{ref:rel18} and Mehta et al. \cite{ref:rel19} that utilize the Partially Observable Markov Decision Processes (POMDP) model to generate navigation control policies. However, unlike CoHAN, no situation assessment is implemented in these works.

In a nutshell, the multi-modality mechanism implemented in CoHAN is the following: 
\begin{enumerate}
    \item The system takes as input the navigation goal;
    \item The \textbf{continuous process} starts:
    \begin{enumerate}
        \item HATEB Local Planner accesses the human-robot scenario and determines the Human State and the Planning State; 
        \item Depending on the value of the determined states, planner shifts between different Planning Modes;
        \item The current Planning Mode is used to choose the command velocity to send to the robot’s base controller;
    \end{enumerate}
    \item The process completes when either the goal is reached, the robot is stuck, or there is a collision, in which case the recovery behavior is activated. 
\end{enumerate}

The available Planning Modes are \verb|SingleBand|, \verb|DualBand|, \verb|VelObs| and \verb|Backoff-recovery|.

\verb|SingleBand| $-$ the mode in which the elastic band is added only to the robot. This mode can be seen as a purely reactive mode. The planning system starts in \texttt{SingleBand} and stays in it as long as there are no humans. This mode is computationally the least expensive.

\verb|DualBand| $-$ the mode in which the elastic bands are added to two nearest observable dynamic humans and to the robot. Trajectories for two humans and a robot are optimized simultaneously. This allows the robot to demonstrate proactive behavior. Additionally, the planned for humans trajectories offer a possible solution for human-robot conflicts.

\verb|VelObs| $-$ the mode in which elastic bands are added to humans, and trajectories for them are predicted only if they have some velocity. This mode is less proactive, but it allows for active re-planning and is useful in crowded scenarios or when the robot cannot move due to the Entanglement Problem of the \texttt{DualBand} mode.  
    
\bigskip
\textit{\textbf{The Entanglement Problem} \cite{ref:hateb}: HATEB assumes that humans keep moving and try to adapt its path according to their motion. This assumption can result in an entanglement of trajectories when the human no longer moves and the robot keeps waiting for the human to move, neglecting the other possible solutions.} 
\bigskip

\verb|Backoff-recovery| $-$ the mode that is activated when the robot encounters the Full Blockage Problem of the \texttt{VelObs} mode. In \texttt{Backoff-recovery}, the robot moves backwards slowly until it finds a free space to clear the way. Once the robot clears the way, it waits for the corresponding human to complete its navigation or a timeout and then proceeds to its goal. 
    
\bigskip
\textit{\textbf{The Full Blockage Problem}: The robot is in the near vicinity of the human, and it is stuck without progressing towards the goal for a time window of a certain duration. There is no solution to the planning problem unless one of the agents completely clears the way for the other. This kind of situation commonly occurs in a very narrow corridor.}
\bigskip

\begin{figure}[h]
    \centering
    \includegraphics[width=\linewidth]{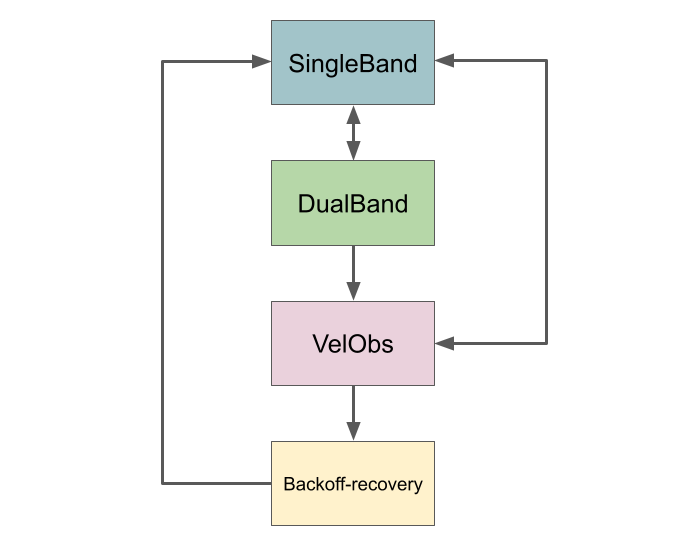}
    \caption{\centering Mode transition scheme. The arrow represents a one-sided transition; the double arrow represents a two-sided transition.}
    \label{fig:modes}
\end{figure}

The decision-making process involved in transitioning between different modes is shown in \autoref{fig:modes}. The shifting criteria are thoroughly explained in \cite{ref:hateb} and \cite{ref:cohan}. On a low level, they are based on a thresholding of the measured human-robot distances and human velocities depending on the states of the observable humans.

\section{Integration}
\label{sec:4}
This section demonstrates how the CoHAN system has been integrated into MARRtina robot software. 

The MARRtina robot software is a \textbf{Docker-based} software. It contains ROS packages representing different robot functionalities and control and visualization modules distributed among Docker images (or, as runtime instances, Docker containers) and interfaced with Python, C++, and other languages. 

There are a number of available \textbf{images}, including the \verb|navigation/navigation-cohan| and \verb|mapping| components responsible for the Navigation Stack functioning: \verb|mapping| contains the Mapping module, and \verb|navigation| stores the executables to perform Localization and to move robot base along the paths provided by the Path Planning. The Path Planning itself is represented by the variety of planning approaches. They are normally implemented as external systems and pulled from the outer repositories into being inherited from the \verb|navigation| image. Similarly, the Co-operative Human-Aware Navigation (CoHAN) planner repository is pulled to the \verb|navigation-cohan| image. Thus, this image becomes a foundation for the overall integration process.

The Docker \textbf{containers} involved in the integration and testing of the CoHAN system are the following:
\begin{itemize}
    \item \verb|navigation|
    \item \verb|base|
    \item \verb|stage|
    \item \verb|nginx|
    \item \verb|xserver|
\end{itemize}

The last two containers in the list are related to the computer set-up (web server); \verb|base| and \verb|navigation| are the runtime instances of the \verb|base| and \verb|navigation-cohan| images, respectively; and \verb|stage| is a simulator.
To launch the navigation, the containers of interest for the execution of the commands are the \verb|stage| and the \verb|navigation|. Hence, in order to integrate and challenge a new planner within the navigation framework, these two containers are the architectural components that must be modified.

The ROS Stage simulator and RViz visualization tool were used to model the navigation scenarios. The Stage\footnote{\url{http://wiki.ros.org/stage}} is a standard ROS 2D simulator, and the corresponding MARRtina software package contains a Python script to automatically create and run the simulated environments with human or robot agents. Mainly, two modifications were made in the \verb|stage| container: 
\begin{enumerate}
    \item The map collection was complemented;
    \item The Human Tracking system was provided.
\end{enumerate}

\begin{figure}[!h]
     \centering
     \begin{subfigure}{.45\linewidth}
         \centering
         \includegraphics[width=.43\linewidth]{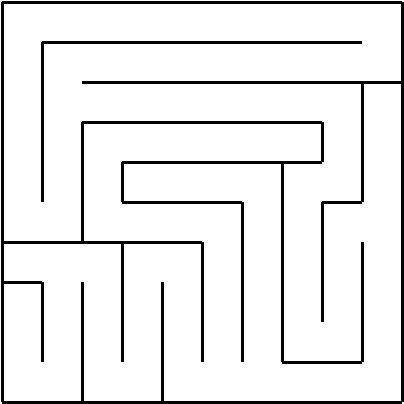}
         \caption{Labyrinth.}
         \label{fig:maze}
     \end{subfigure}
     \begin{subfigure}{.85\linewidth}
         \centering
         \includegraphics[width=0.43\linewidth]{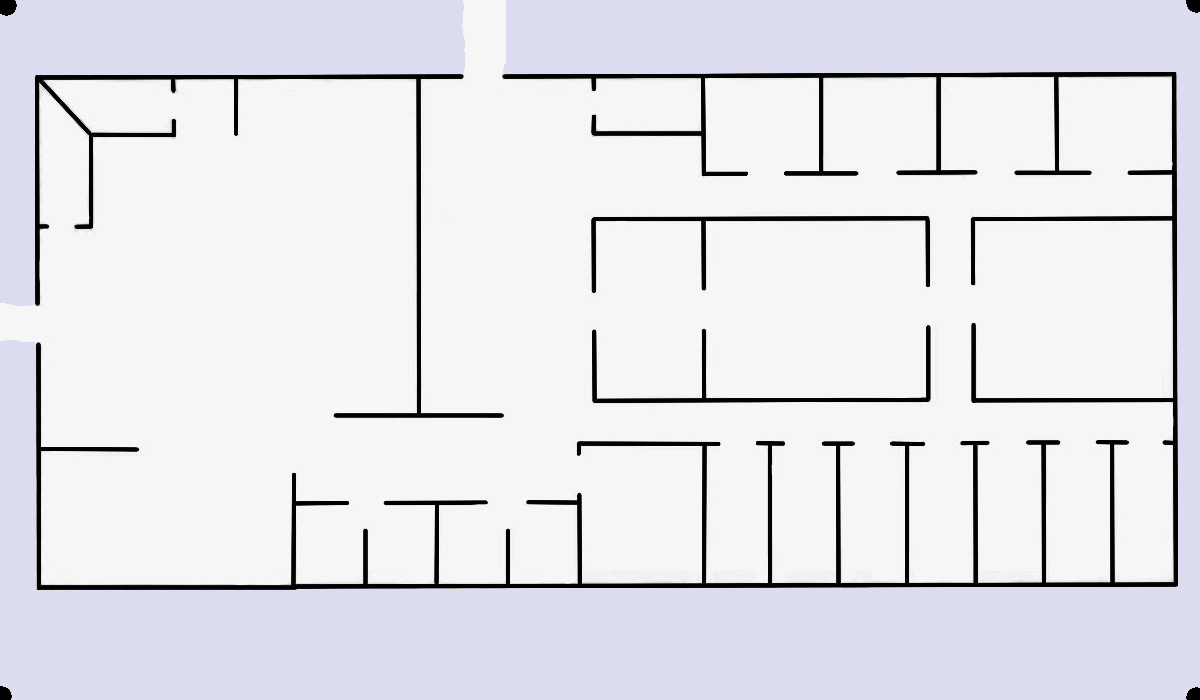}
         \caption{Emergency Department.}
         \label{fig:er}
     \end{subfigure}
     \begin{subfigure}{.45\linewidth}
         \centering
         \includegraphics[width=.43\linewidth]{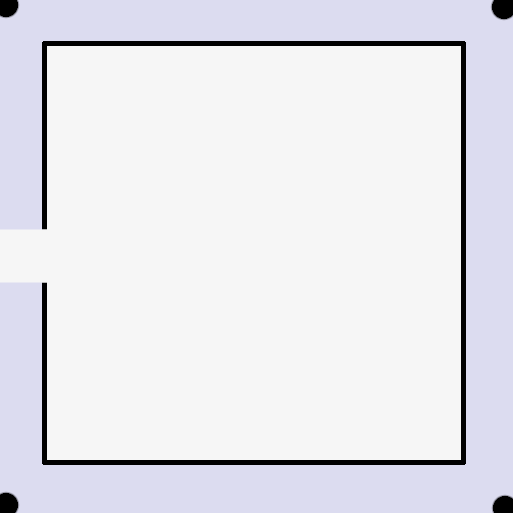}
         \caption{ICU.}
         \label{fig:icu}
     \end{subfigure}
     \begin{subfigure}{.45\linewidth}
         \centering
         \includegraphics[width=.43\linewidth]{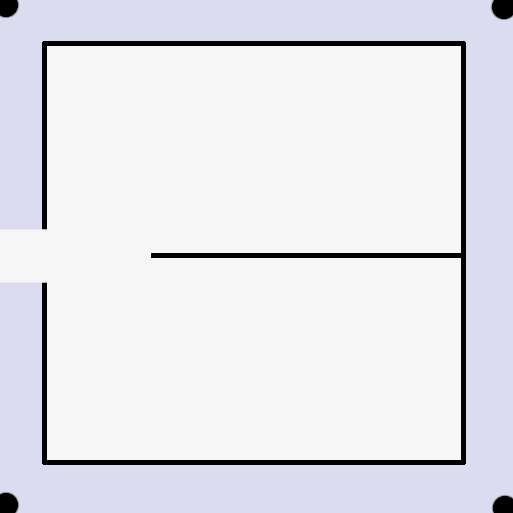}
         \caption{ICU2.}
         \label{fig:ICU2}
     \end{subfigure}
        \caption{\centering Map files used to simulate the stage.}
        \label{fig:maps}
\end{figure}

First, some of the existing maps were customized and/or augmented with semantic information, while others were created from scratch because a large collection of maps is required to conduct extensive testing. On the one hand, the task was to model all of the intricate scenarios that the planner was originally designed to handle efficiently (e.g., doorways, hallways, and corridors) in order to verify its declared properties. On the other hand, since enhancement of a robot's reliability in healthcare environments is our focus, the aim was to learn how to account for the specificity of a domain the best by simulating medical contexts. \autoref{fig:maps} shows four main maps used in experiments: labyrinth, Emergency Department, Intensive Care Unit (ICU), and Intensive Care Unit with the wall in the middle (ICU2). 

Second, as we remember from the \hyperref[subsec:tracking]{Section~\ref*{subsec:tracking}} and \autoref{fig:architecture}, there is an external system in charge of Human Tracking. According to its arrangement, to include humans into the system, they must be published on a \verb|/tracked_humans| topic following the particular message structure. This is implemented in a Python bridge script that takes as input the number of humans and then pipes a subscriber to the humans and a publisher to \verb|/tracked_humans|. The script must have been added to the \verb|stage| to function properly.

After Human Tracking is activated and the Localization (in this case, basic \verb|amcl|) is launched, the core component of the Navigation Stack must be started. This main executable is called \verb|cohan_nav.launch|. It was complemented with the necessary frame transformations and placed in the \verb|navigation|, together with new configuration files containing \verb|move_base|, Local Costmap, Global Costmap, and HATEB Local Planner's parameters. The \verb|cohan_nav.launch| uses HATEB as a Local Planner in the \verb|move_base| node and, besides this node, contains Human Path Prediction and Human Filtering.

\section{Experiments}
\label{sec:5}
Let us move on to the experiments that have been conducted to create a complete characterization of the integrated system and assert enhancement of the robot's reliability by means of a chosen Human-Aware Navigation Planning approach.

The results in various simulated human-robot contexts are presented and thoroughly analyzed qualitatively and quantitatively, followed by a statistical analysis of the system evaluation by clinicians in terms of acceptability and usability in a real environment. Precisely, the following scenarios have been modeled and evaluated:

\begin{itemize}
    \item \textbf{Visibility Test}
    \begin{itemize}
        \item A human in open space
        \item A human and a wall
        \item Two humans in open space
    \end{itemize}
    \item \textbf{Door Crossing Scenario}
    \begin{itemize}
        \item Wide Doorway
        \item Narrow Doorway
        \item Bed Approach
    \end{itemize}
    \item \textbf{Narrow Corridor Scenario}
    \begin{itemize}
        \item A "never stopping human"
        \item A human who stops
    \end{itemize}
    \item \textbf{Wide Corridor Scenario}
    \begin{itemize}
        \item Free Corridor
        \item Cluttered Corridor
    \end{itemize}
    \item \textbf{Crowded Scenario}
    \begin{itemize}
        \item Free Space Crowd
        \item Emergency Situation
    \end{itemize}
\end{itemize}

The \verb|rqt_reconfigure| ROS plugin was employed to tune the navigation parameters at runtime.

\subsection{Qualitative Results}
\label{subsec:qualitative}
In all the snapshots from RViz, dark and light blue lines are the global path and elastic band (local path) of the robot, dark and light green lines are the global paths and elastic bands generated for humans, yellow arrows indicate human orientations, and the trajectories of the robot and humans are shown as colored bars or dots. They are the poses planned by HATEB Local Planner, and the color visualizes the timestamp. If the color of the predicted human pose bar is the same as the color of the robot pose bar, then they will both be at that location at the same time.

\subsubsection{Visibility Test}
\label{subsubsec:visibility}
This test considers the presence of only static humans and is intended to estimate the influence of a particular component of the system: the Human Safety and Human Visibility layers.

\begin{figure}[!h]
     \centering
     \begin{subfigure}{0.49\linewidth}
         \centering
         \includegraphics[width=\linewidth]{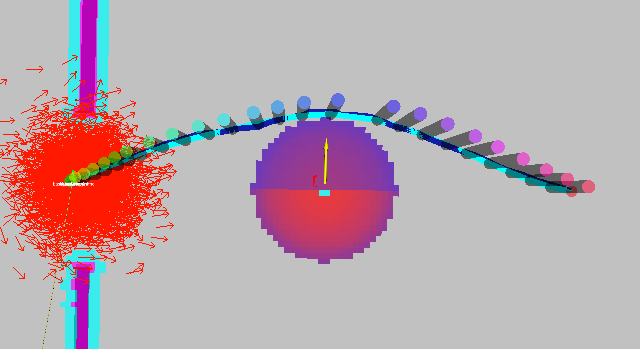}
         \caption{\centering The robot passes in front.}
         \label{fig:vis11}
     \end{subfigure}
     \begin{subfigure}{0.49\linewidth}
         \centering
         \includegraphics[width=\linewidth]{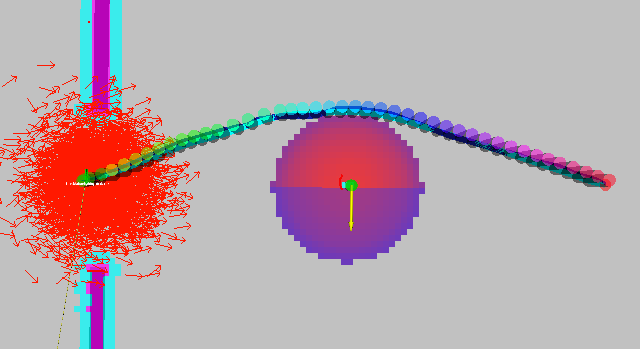}
         \caption{\centering The robot passes behind.}
         \label{fig:vis12}
     \end{subfigure}
        \caption{\textbf{Visibility Test:} A static human in open space. RViz.}
        \label{fig:vis1}
\end{figure}

We start with a test of a single static human in an open space. Since the human is static, the robot is always in \verb|SingleBand|, and the system adds the \verb|human_layers| to the costmaps. As it can be seen from \autoref{fig:vis1}, sometimes robot chooses to pass in front (\autoref{fig:vis11}) and sometimes $-$ behind (\autoref{fig:vis12}) the human. This is explained by the fact that the cost of passing behind or in front is the same outside the safety radius (1.5 m), and since the test is performed in the open space, there are cells of the same cost from both sides. In other words, this is a costmap-based implementation, and the situation is symmetric. The robot has enough space behind the human so as not to disturb him or her and still continue to its goal. The planner does not always make the robot choose the frontal side always $-$ only when the back side is constrained and the robot has to move close to human. However, even when the robot appears from behind the human, it enters the human’s field of view from a larger distance.

\begin{figure}[h]
    \centering
    \includegraphics[width=.6\linewidth]{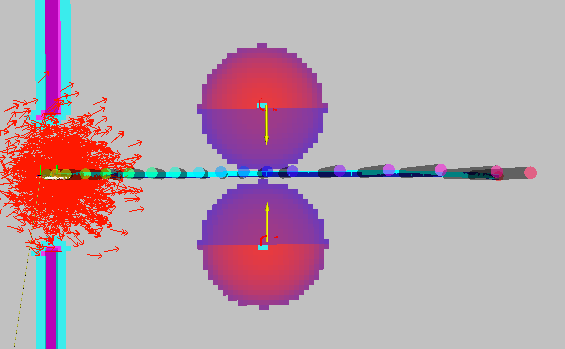}
    \caption{\centering \textbf{Visibility Test:} There is a free space between two static contr-directed humans in an open space that the robot chooses to pass through. RViz.}
    \label{fig:vis32}
\end{figure}

The explanation is confirmed in tests with one static human standing next to the wall and two static humans standing next to each other in an open-space environment. In the case of a static human and a wall, regardless of human orientation, if there is a free space between a human and a wall, the robot navigates through this space; otherwise it navigates around the human. When there are two humans next to each other in open space, no matter if they are co-directed or contr-directed (in which case they can be considered interacting), if there is a free space between them, the robot moves through this space; otherwise, it goes around. An interesting situation is shown in \autoref{fig:vis32}. Usually in Social Navigation, when two humans are facing each other, the robot should not pass between them (interaction should not be interrupted), but from the other side, the distance between these humans is more than 3m.

\subsubsection{Door Crossing Scenario}
\label{subsubsec:door}
The Door Crossing scenario is a series of three simulated situations: Door Crossing: Wide Doorway, Door Crossing: Narrow Doorway, and a Bed Approach Test.

The first two are common situations in many human environments, including medical institutions, because they can occur at the entrance of any room. A dynamic human is simulated as moving with a constant linear velocity along a straight line, with a goal somewhere on this line behind the initial position of the robot. The \verb|PredictBehind| Human Path Prediction service is activated. 

In a Door Crossing: Wide Doorway case a doorway is wide enough for two agents. This scenario is appropriate to test the robot in a \verb|DualBand| mode, because there is a dynamic human, and neither Entanglement, nor Blockage Problem can happen. The test completes successfully: robot avoids collisions and demonstrates the declared behavior. The robot always correctly switches from a \verb|SingleBand| to a \verb|DualBand|, but the other events vary depending on the run. 

\begin{figure}[h]
    \centering
    \includegraphics[width=.75\linewidth]{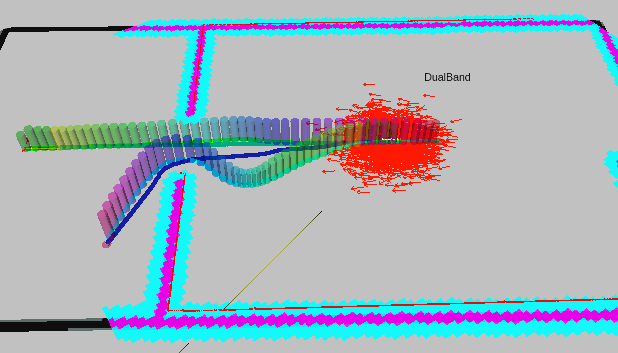}
    \caption{\centering \textbf{Door Crossing:} Wide Doorway. Perfect run. The robot's elastic band suggests more effort for a robot. RViz.}
    \label{fig:dcperfect}
\end{figure}

During the perfect run, illustrated in \autoref{fig:dcperfect}, the robot is able to guess the linear motion of the human and plan its own path with a deviation maneuver to let the human pass first. In this case, the robot makes a greater effort than a human.

In the case of a less lucky run, the robot does not plan a greater effort for itself because it assumes human cooperation (i.e., that human will move aside too). Fortunately, this does not cause collisions since the robot finds an alternative to the waiting solution: to pass through the cell of a doorway that is not occupied by the human.

\begin{figure}[!h]
    \centering
    \includegraphics[width=.75\linewidth]{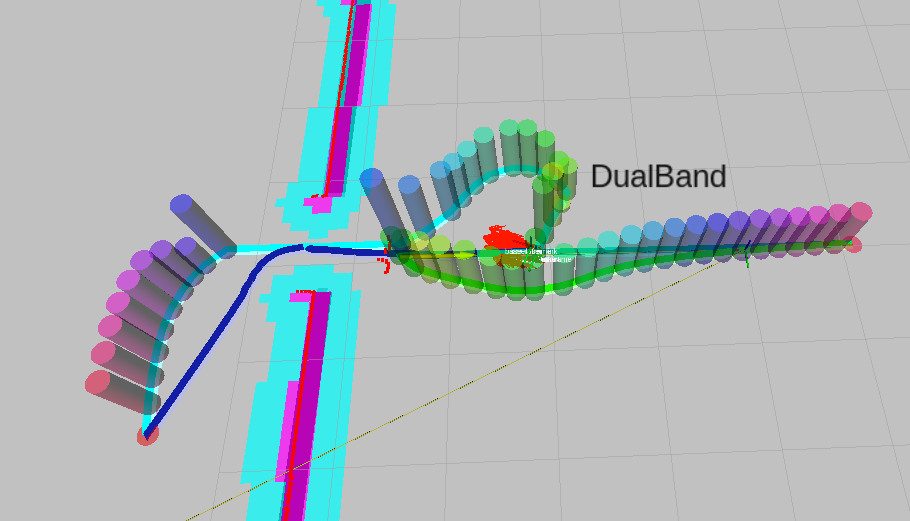}
    \caption{\centering \textbf{Door Crossing:} Narrow Doorway. Human is assumed to be cooperative, but the robot backtracks. RViz.}
    \label{fig:dc2}
\end{figure}

A similar problem is present in the Door Crossing: Narrow Doorway scenario, where the setup is the same but the doorway is wide enough for one agent only. The tests for this scenario were also successful. However, this task is more complicated because simultaneous crossing is not possible and the process of entering the doorway requires higher precision. As a result, the robot does not ever guess a human's linear motion and does not plan maneuvers to move aside. Instead, it again assumes human cooperation and hence proposes a cooperative solution. Then, at the moment right before the collision, shown in \autoref{fig:dc2}, the robot takes an action for avoidance and starts moving backwards to let human pass. The robot keeps backtracking until the doorway is free. Eventually, it re-plans and proceeds to the goal in a \verb|SingleBand|.

The robot's planner recommends the cooperative solution in both Door Crossing scenarios, partly due to inaccuracy in the human model but primarily due to this being a property of CoHAN by design: it assumes that both agents are interested in cooperation. The system does not predict the human's path but rather plans it based on the estimated goal and the human model. This is a proactive planning approach. It is not very accurate, but the robot's behavior is enhanced in comparison with non-social planners. The elimination of these inconsistencies points to directions for future work, such as improving goal prediction and updating the human model.

The series of experiments performed to test robot capabilities in a human-robot collaborative crossing context ends with a scenario that is especially relevant in the healthcare sector: the Bed Approach scenario. The map of an Intensive Care Unit with a wall in the middle of the room and beds placed along the walls is used. The Door Crossing situation is modeled by two agents conflicting for exiting/entering the free space between two hospital beds: the human provider is exiting and the robot is entering. The goals of both agents are behind each other. Again, the \verb|PredictBehind| service is set on. The focus is not on the quality of the approach itself but on the behavior of the robot towards human.

\begin{figure}[!h]
     \centering
     \begin{subfigure}{.65\linewidth}
         \centering
         \includegraphics[width=\linewidth]{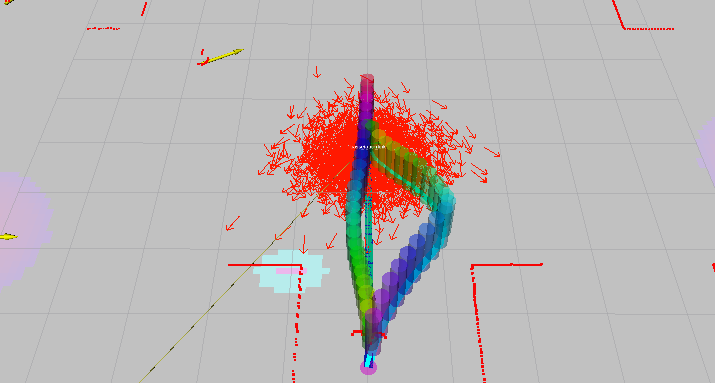}
         \caption{\centering A red point cloud denotes the robot's location guesses. RViz.}
         \label{fig:ba1}
     \end{subfigure}
     \begin{subfigure}{.65\linewidth}
         \centering
         \includegraphics[width=\linewidth]{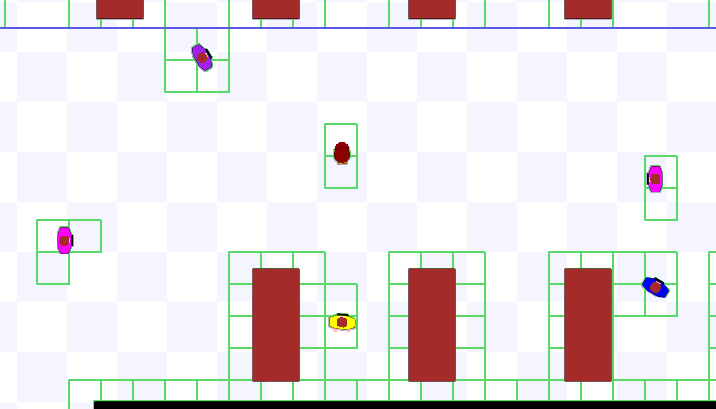}
         \caption{\centering A robot is depicted in dark red, and a dynamic human is yellow. Stage.}
         \label{fig:ba2}
     \end{subfigure}
        \caption{\centering \textbf{Door Crossing:} Bed Approach. The robot's elastic band suggests more effort for a robot.}
        \label{fig:ba}
\end{figure}

The test is successful because the robot collision never happens, and, normally, human collision does not happen either because the robot demonstrates the expected behavior:
\begin{enumerate}
    \item The human and the robot move towards each other;
    \item As soon as the robot detects the human, it activates \verb|DualBand| mode and derives path predictions for itself and for the human while imposing more effort on itself (\autoref{fig:ba});
    \item The more human-robot distance reduces, the more the robot slows down;
    \item When the distance becomes critically small, the robot moves a bit backwards, stops, and remains in this state until the human passes it by;
    \item The robot resumes to its goal.
\end{enumerate}

\subsubsection{Narrow Corridor Scenario}
\label{subsubsec:narrow}
The Narrow Corridor Scenario can happen in any hospital corridor, and it challenges the planner with the Blockage Problem: a long corridor has to be traversed by two agents in opposite directions, and the corridor is wide enough only for a single agent; one of the agents must go back and wait for the other to cross. When one of the agents is a robot, it should back off, giving priority to the human while taking legible actions. Thus, the expected behavior of the robot is the following: 
 \begin{enumerate}
     \item The robot switches from \verb|SingleBand| to \verb|DualBand| and plans a certain trajectory;
     \item The robot’s way is blocked by the human, and the system shifts to the \verb|Backoff-recovery| mode (the \verb|Backoff-recovery| is supposed to be activated when the robot is in \verb|VelObs| mode in the near vicinity of the human (<2.5m), and it is stuck without progressing towards the goal);
     \item The robot clears the way for the human, drives away from the corridor, and waits on the side until the human crosses the robot;
     \item The robot resumes to its goal in \verb|SingleBand| mode.
 \end{enumerate} 
 
The dynamic human is moving with a constant linear velocity along a straight line and is a non-collaborative human (does not move back). We consider two different approaches to modeling this human's behavior: one is the human who never stops, and, alternatively, the human who moves but then stops. Both variations are interesting because they highlight certain features of the robot's behavior. 

A "never-stopping human" keeps moving in the corridor even if there is a robot on the way. In real life, in most situations, human will not be "never-stopping". However, this situation can occur, for example, when a person is looking at the phone and hence does not see that there is a robot in the corridor.
 
\begin{figure}[!h]
    \centering
    \includegraphics[width=.75\linewidth]{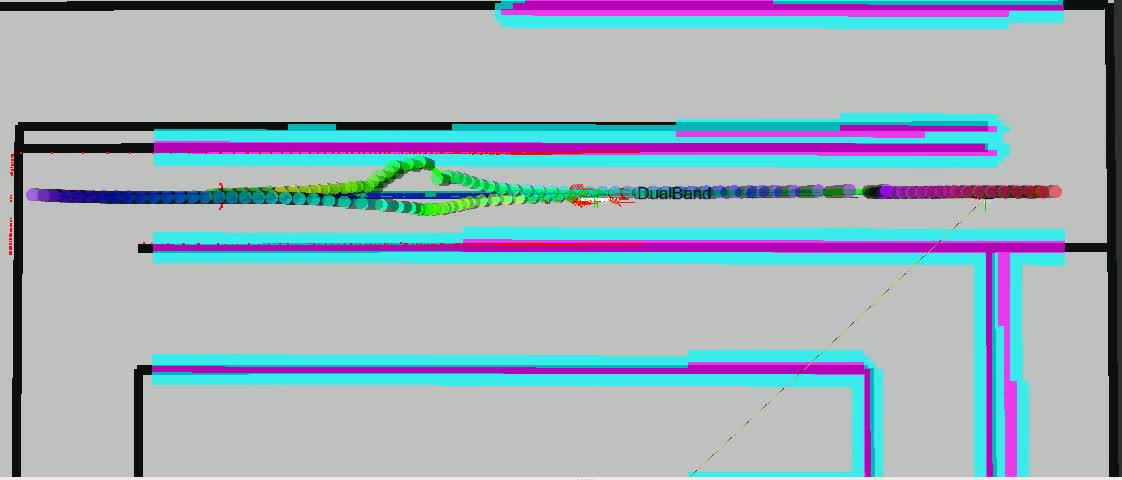}
    \caption{\centering \textbf{Narrow Corridor:} A "never-stopping human". The robot remains in \texttt{DualBand} after meeting human. RViz.}
    \label{fig:nc1}
\end{figure}

The test always results in a human collision. Right before the collision, the robot slows down and attempts to move backwards, but at some point it abandons its attempts and takes a hit. This happens because the robot stays in \verb|DualBand| mode ever since it detects a dynamic human (\autoref{fig:nc1}): the condition for a shift to \verb|VelObs| $-$ the human stopping $-$ is never satisfied. In turn, without shifting to \verb|VelObs|, a shift to \verb|Backoff-recovery| cannot be performed as well. The negative outcome is not critical because the collision is, in fact, a human's fault. Yet, the test case demonstrates an important limitation of the CoHAN planner.

 In another variation of the scenario, some time after the robot detects the dynamic human, the human fully stops. In this case, the expected behavior is not confirmed through experimentation either. The robot gets stuck either without a shift \verb|VelObs| $\rightarrow$ \verb|Backoff-recovery| (a shift \verb|DualBand| $\rightarrow$ \verb|VelObs| is performed correctly) or with a shift, but without going backwards. The problem here is that the functioning of the robot in \verb|Backoff-recovery| mode is prone to errors because the mode implementation lacks robustness. This is regarded as one of the algorithm's current limitations that leaves room for improvement. For the time being, the result of the robot stopping in the narrow corridor after detecting the Blockage is considered satisfactory as opposed to the collision.
 
\subsubsection{Wide Corridor Scenario}
\label{subsubsec:wide}
\begin{figure}[!h]
    \centering
    \includegraphics[width=.85\linewidth]{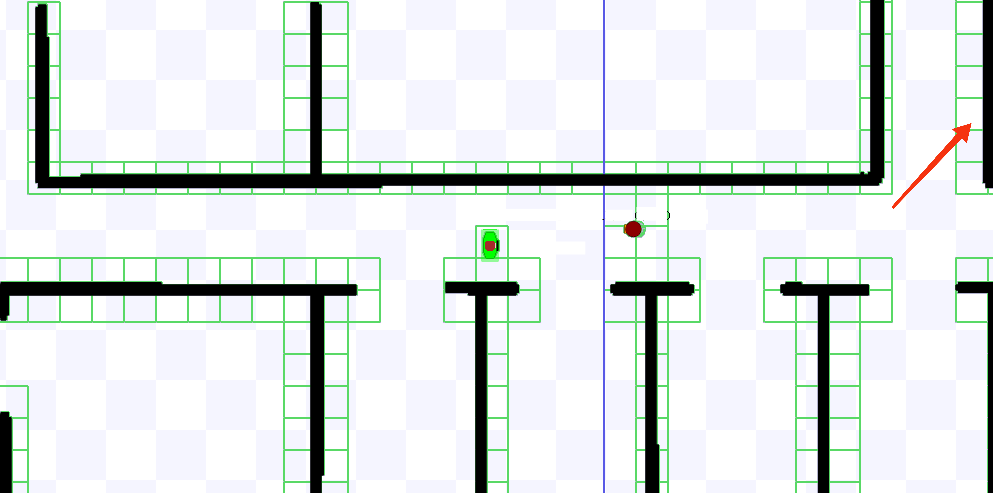}
    \caption{\centering \textbf{Wide Corridor:} Free corridor. Initial setup, where the human goal is denoted by the red arrow. Stage.}
    \label{fig:wcstage}
\end{figure}

The test scenario takes after the Narrow Corridor Scenario, but now the corridor is large enough for two agents. Additionally, on one side of the corridor there are rooms with open doors, so it is possible for the robot to drive there and wait until a human passes. The human goal is indicated in \autoref{fig:wcstage}. Since the human intends to turn, the human agent's motion is nonlinear. The \verb|PredictGoal| Human Path Prediction service is activated.

Once again, there are two modifications of the scenario: Wide Corridor: Free Corridor and Wide Corridor: Cluttered Corridor.

\begin{figure}[!h]
    \centering
    \includegraphics[width=.75\linewidth]{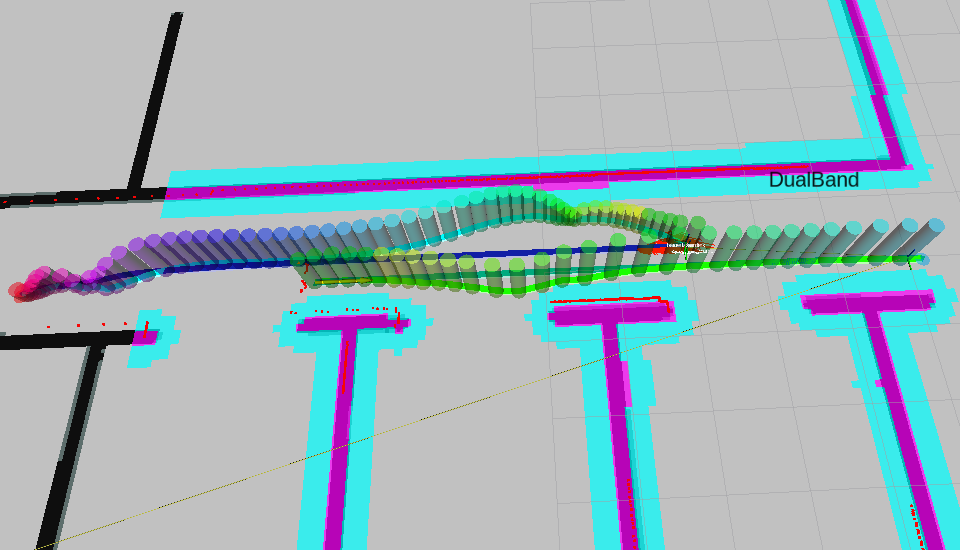}
    \caption{\centering \textbf{Wide Corridor:} Free Corridor. Perfect run. The robot's elastic band suggests more effort for a robot. RViz.}
    \label{fig:wc1}
\end{figure}

\begin{figure}[!h]
    \centering
    \includegraphics[width=.75\linewidth]{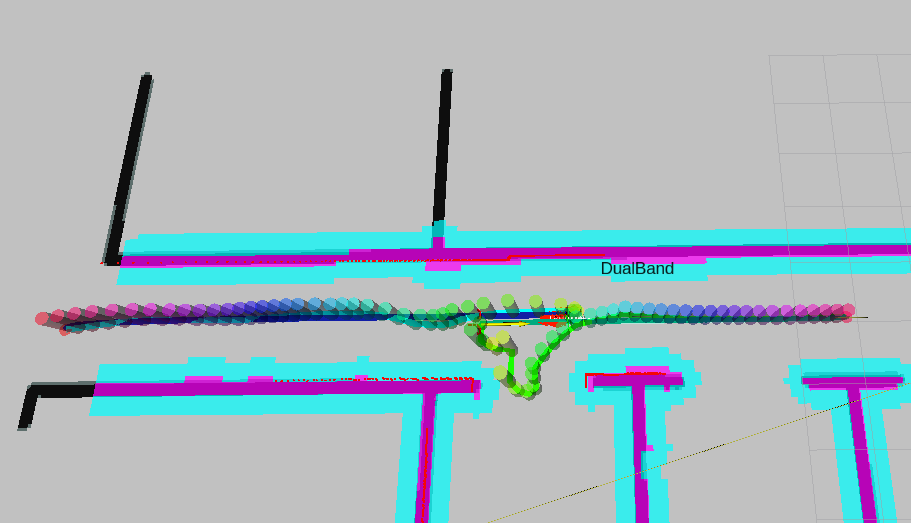}
    \caption{\centering \textbf{Wide Corridor:} Free Corridor. The human's elastic band suggests justified extra effort for a human. RViz.}
    \label{fig:wc2}
\end{figure}

The Wide Corridor: Free Corridor challenge is usually resolved positively by the robot; the collision does not occur and the mode transitions are performed properly. In a perfect run, the cooperative solution that suggests a greater effort for a robot is planned (\autoref{fig:wc1}).
Otherwise, the cooperative solution proposes a heavier load for a human. But in the majority of cases this is justified, as in the \autoref{fig:wc2}, for example: the moment is close to a collision and the robot's current velocity does not allow it to perform the maneuver of entering the free room doorway, so the robot expects this maneuver to be executed by the human. However, the system's capacity to anticipate navigation conflicts requires improvement because the situations illustrated in \autoref{fig:wc2} can be prevented if the robot foresees them. Additionally, the possibility of driving off is completely ignored by the robot. Presumably, Human Path Prediction is a responsible module of architecture that needs to be upgraded.

Moving on to a Wide Corridor: Cluttered Corridor, the setting is similar, but now a more realistic clinical environment is simulated. As it is investigated in \cite{ref:situating}, patients are often treated in the corridors when the Emergency Department (ED) bedrooms are full. Placing patients in hallways is a way to handle an overflow of patients. Therefore, the corridors are often overcrowded and cluttered with stretchers and other medical equipment.

We model a simplified version of the Cluttered Corridor scenario where there is only one dynamic human and all the stretchers are leaned against the same wall. Only one agent at a time can pass between the stretcher and the other wall. The robot and the human start navigating from the opposite sides of the cluttered part of the corridor, and their goals are behind each other. 

The test normally produces the same series of events at each run: 
\begin{enumerate}
    \item Following its initial path, the robot arrives in the middle of the cluttered hallway, where it detects a human and switches from \verb|SingleBand| to \verb|DualBand|;
    \item The robot backs off all the way back until it reaches the vicinity of its initial position and then comes to a halt.
\end{enumerate} 

\begin{figure}[!h]
    \centering
    \includegraphics[width=.75\linewidth]{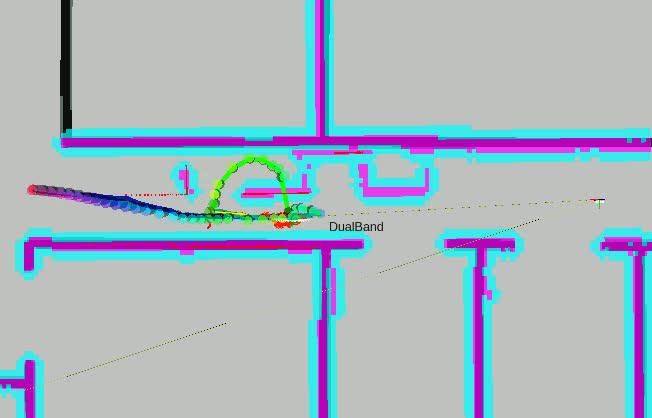}
    \caption{\centering \textbf{Wide Corridor:} Cluttered Corridor. The human's elastic band suggests extra effort for a human. RViz.}
    \label{fig:cc}
\end{figure}

This test result leaves a dual impression. The robot exhibits safe behavior that enables collision avoidance. It even gives priority to a human and does not insist on pursuing its goal; it also does not abort in the middle of the corridor but moves backward and clears the way instead. Unfortunately, the robot is neither proactive nor anticipative enough. It is not anticipative due to the fact that it only begins to resolve a conflict after it has already encountered it, not in advance. And non-proactivity is well-seen in the \autoref{fig:cc}, where there are three options for one agent to clear the way to the other: two free spaces between the stretchers and one free room entrance. The CoHAN system proposes a deviating elastic band for a human. In \autoref{fig:cc}, the deviation maneuver opts for the first (counting from left) free space between the stretchers, but as simulation proceeds, it changes to the second and third ones, respectively (always for a human).

\subsubsection{Crowded Scenario}
\label{subsubsec:crowded}
The series of experiments concludes with a Crowded Scenario, which is the most challenging for the CoHAN system. The scenario consists of two sub-scenarios invented to estimate the robot's reliability in a crowd from two different perspectives. Both scenarios are simulated in an Intensive Care Unit (ICU) environment.

\begin{figure}[!h]
    \centering
    \includegraphics[width=\linewidth]{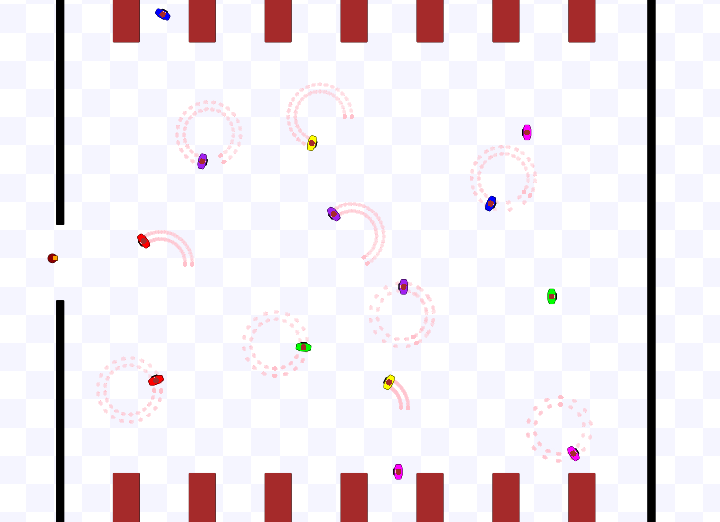}
    \caption{\centering \textbf{Crowded Scenario:} Free Space Crowd. Initial setup. Humans' footprints represent their motion. Stage.}
    \label{fig:cs}
\end{figure}

The first sub-scenario covers the context of a crowd in the free space and is named Free Space Crowd. The map is a room that does not include any static semantic objects other than the beds along the walls. The humans are distributed around the room and form a chaotic crowd. There are 14 humans in total: 10 of them are dynamic and 4 are static. The dynamic humans have different constant velocities and move along a circular trajectory each (\autoref{fig:cs}). The robot's objective is to visit all of the static humans while navigating safely and intelligently through the dynamic crowd.

\begin{figure}[!h]
    \centering
    \includegraphics[width=.75\linewidth]{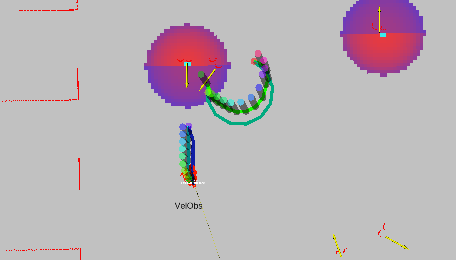}
    \caption{\centering \textbf{Crowded Scenario:} Free Space Crowd. The robot in \texttt{VelObs} is able to predict the circular trajectory. RViz.}
    \label{fig:cs1}
\end{figure}

The most evident positive aspect of navigation in the crowd can be observed from the \autoref{fig:cs1}: the robot in \verb|VelObs| is able to correctly predict the humans' paths. In fact, \verb|VelObs| is the robustest Planning Mode of CoHAN.

\begin{figure}[!h]
    \centering
    \includegraphics[width=.75\linewidth]{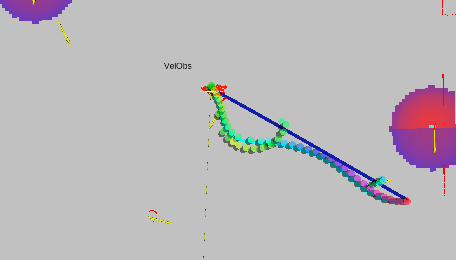}
    \caption{\centering \textbf{Crowded Scenario:} Free Space Crowd. The robot adapts to the human. RViz.}
    \label{fig:cs2}
\end{figure}

\begin{figure}[!h]
    \centering
    \includegraphics[width=.75\linewidth]{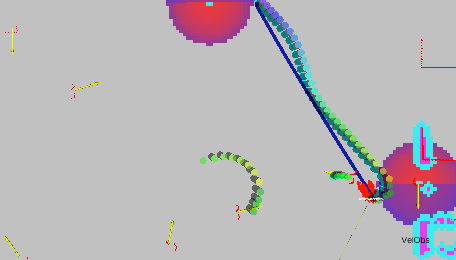}
    \caption{\centering \textbf{Crowded Scenario:} Free Space Crowd. The temporary blockage resolution by oscillation. RViz.}
    \label{fig:cs3}
\end{figure}

Two other remarkable examples of intelligent robot behavior are shown in \autoref{fig:cs2} and \autoref{fig:cs3}. In \autoref{fig:cs2}, the human with a lower velocity emerges on the robot's path. The robot slows down (or stops if necessary) and adapts its velocity. Then, it starts following the human from behind until its path no longer passes through the risk zone. The social nuance is whether the human feels comfortable being followed.

In \autoref{fig:cs3}, the robot is confronted with a situation in which it is trapped between a static human and inflated static obstacles on one side and a human moving extremely slowly on the other side. The slow human temporarily blocks the robot. The robot resolves the temporary blockage through oscillation. Every time it moves forward, the robot checks if the person has moved away; if not, the robot continues to oscillate; otherwise, it proceeds to the goal.

The second sub-scenario models the situation of emergency in the ICU: the patient needs urgent intervention from the providers, e.g., he or she has a heart attack. In such circumstances, the robot's proactivity is especially important because it must not interrupt providers performing life-saving treatment on a patient. However, the desired robot behavior is difficult to provide because the robot bases its reasoning on a certain system of beliefs about the state of the environment, and this state changes abruptly in emergencies.

\begin{figure}[!h]
    \centering
    \includegraphics[width=.9\linewidth]{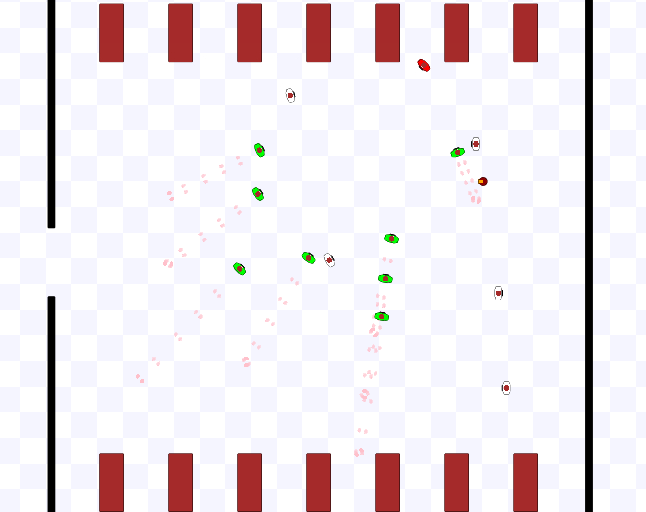}
    \caption{\centering \textbf{Crowded Scenario:} Emergency Situation. Initial setup. Humans' footprints represent their motion. There are 14 people in the environment, divided into three categories: \textbf{1 red human} $-$ "emergency" $-$ location of emergency; \textbf{8 green humans} $-$ "help" $-$ providers who rush to the location of emergency; \textbf{5 white humans} $-$ "neutral" $-$ providers who cannot abandon their current tasks and remain at the initial positions in the map. Stage.}
    \label{fig:hastage}
\end{figure}

In the simulation, an emergency is occurring at a specific location: one of the providers is standing next to the bed of the patient with the emergency, and he or she is calling other providers to help. The scenario assumes that 10 providers suddenly start running to the provider who is calling. They form a "running crowd" next to the bed of the patient with an emergency (\autoref{fig:hastage}). Meanwhile, the robot had been following a path to some goal, but this path now passes through the emergency zone (i.e., comes into conflict with the paths of the "running crowd"). The robot's expected behavior is to avoid interfering with the crowd.  

The robot operates in \verb|VelObs| whenever dynamic humans enter the area, limited by the Planning Radius. The robot collision never occurs. However, a human collision may occur, depending on the test run.

\begin{figure}[!h]
    \centering
    \includegraphics[width=.75\linewidth]{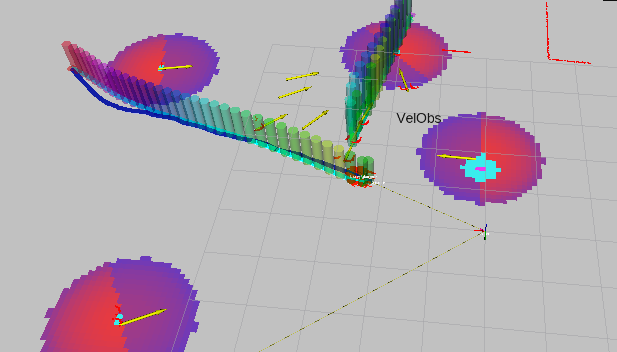}
    \caption{\centering \textbf{Crowded Scenario:} Emergency Situation. Lucky run. The robot passes around the "running crowd." RViz.}
    \label{fig:ha1}
\end{figure}

\begin{figure}[!h]
    \centering
    \includegraphics[width=.75\linewidth]{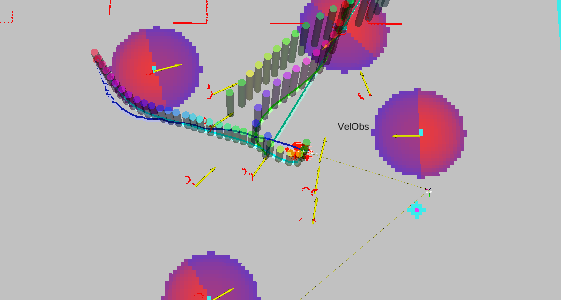}
    \caption{\centering \textbf{Crowded Scenario:} Emergency Situation. Lucky run. The robot was able to find a "safe spot." RViz.}
    \label{fig:ha2}
\end{figure}

In lucky runs, the robot is able to re-plan and avoid safely either by passing around the crowd, as presented in the \autoref{fig:ha1}, or by following its path while adapting the speed until the "safe spot" is found and halting at this spot to wait for the crowd to move away (\autoref{fig:ha2}).

When the robot is less lucky, it gets trapped by humans with no "safe spot" found. In this case, a human collision happens: the robot slows down and comes to a halt at the point where it realizes there is no "safe spot," and one or more running humans collide with it. The issue once again uncovers a point for improvement in the robot's inability to be anticipative enough to foresee the human collision and take look-ahead action. Additionally, it is linked to the fact that the robot derives plan predictions for the two nearest dynamic humans only, and in emergencies, more than two of them are relevant.

This final test concludes the section on qualitative results.

\subsection{Quantitative Results}
\label{subsec:quantitative}
Five experimental scenarios have been chosen to perform the quantitative analysis, each repeated 10 times with the Co-operative Human-Aware Navigation (CoHAN) and Simple Move Base (SMB) systems in order to compare an integrated Social Navigation planner to a non-human-aware approach such as those found in MARRtina software prior to integration. The averaged over 10 runs results are collected in the \autoref{table:quantitative}.

The tested scenarios are Door Crossing: Bed Approach, Narrow Corridor: move-and-stop human, Wide Corridor: Free Corridor, Crowded Scenario: Free Space Crowd, and Crowded Scenario: Emergency Situation. In all these scenarios, both systems produced consistent results over repeated trials. However, in two of them, SMB failed to complete collision-free navigation.

\begin{table*}[!htb]
\centering
\caption{\centering \textbf{Mean values of the metrics over 10 repetitions in five different scenarios:} Acc $-$ Accuracy, PL $-$ Path Length, TT $-$ Total Time, HRD $-$ Minimum Human-Robot Distance (0 if at least once the collision happened). The test failure is denoted by "-."}
\label{table:quantitative}
\begin{tabular}{ |l|c|c|c|c|c|c|c|c|  }
 \hline
  \multirow{2}{*}{Scenario} & \multicolumn{4}{c|}{CoHAN} & \multicolumn{4}{c|}{SMB} \\ \cline{2-9}
   & Acc[\%] & PL[m] & TT[s] & HRD[m] & Acc[\%] & PL[m] & TT[s] & HRD[m] \\
 \hline
 \textit{Bed Approach} & \textbf{76} & 5.07 & 13.04 & 0.51 & - & \textbf{4.34} & - & 0 \\
 \hline
\textit{Narrow Corridor} & 50 & 31.12 & 93.99 & 1.19 & \textbf{100} & \textbf{8.92} & \textbf{34.52} & \textbf{2.98} \\
 \hline
 \textit{Wide Corridor} & 100 & 16.73 & \textbf{27.66} & \textbf{0.89} & 100 & \textbf{16.22} & 32.02 & 0.50\\
 \hline
 \textit{Free Space Crowd} & \textbf{72} & 71.66 & 155.06 & 0 & 50 &  \textbf{55.51} & \textbf{138.46} & 0 \\
 \hline
 \textit{Emergency Situation} & \textbf{50} & 16.98 & 43.78 & 0 & - & \textbf{9.65} & - & 0 \\
 \hline
\end{tabular}
\end{table*}

The metrics involved in comparison are the accuracy (Acc), the total length of the path (PL), the total execution time (TT), and the minimum human-robot distance that the planner encountered during navigation (HRD).

Accuracy is the percentage of times a test is completed successfully. "Success" is defined as reaching the navigation goal without colliding and getting stuck, except for the Narrow Corridor, where it is redefined as a detection of the Blockage followed by abortion. This always happens when the robot is controlled with a Simple Move Base, but CoHAN distinguishes between the robot aborting and getting stuck. The navigation cannot be resumed in the latter case, and this occurs $50\%$ of the time. In all the scenarios, besides the corridors, CoHAN outperforms SMB in accuracy. In fact, SMB does not avoid human collisions at all in the Bed Approach and Emergency Situation scenarios, whereas using a new planner robot sometimes gets lost in the bed inflation or is not able to find a "safe spot," but sometimes resolves the conflict. In the Free Space Crowd, the robot with SMB can only navigate safely when it is lucky. The CoHAN system, on the other hand, produces rare collisions due to lag in costmap updates during mode switching. Finally, both approaches are reliable in the free wide corridors.

The path length is the total length of the path, computed as the sum of the distances between every sequential pair of states. When SMB tests fail, the length of the generated global path is computed instead. The \autoref{table:quantitative} shows that in all test instances, SMB made robot travel shorter distances. This is explained by HATEB using larger deviations for early intention show, proactive elastic bands, and driving away maneuvers. Nonetheless, there is a specific reason for the large difference in path lengths in the Narrow Corridor scenario: the non-socially navigating robot comes to a halt as soon as it detects a blocking obstacle, and the CoHAN-based system, before declaring abortion or understanding that the robot is stuck, spends some time on mode switching and subsequent re-planning while oscillating. The same reason explains why CoHAN takes nearly three times as long as SMB to complete the navigation in a given scenario and why the minimum human-robot distance is more than two times shorter.

Due to proactive deviation maneuvers performed by CoHAN, the total execution time taken by it is greater than that of SMB unless the human-robot co-navigation context is such that the reactive planner takes longer because the robot needs to slow down in the human vicinity for collision avoidance, as happens in Wide Corridor experiments. 

Finally, when it comes to the minimum human-to-robot distances, the HATEB's behavior varies because it performs the situation assessment and addresses each case individually. If the space is available, the integrated planner keeps a greater minimum distance from the human than SMB. Otherwise, it can also choose the strategy of slowing down and approaching closer.

In summary of the quantitative evaluation, the integrated system proves its human awareness and can be considered outperforming in all simulated scenarios, except for the Narrow Corridor, where the Simple Move Base exhibits more reasonable and less resource-consuming behavior. However, even in the latter case, the conceptual difference is not large and can be mitigated by debugging the \verb|Backoff-recovery| mode.

\subsection{Human Evaluation Results}
\label{subsec:acceptability}

In order to assess human acceptance of the assistive robot driven by the integrated system and estimate its usability in health-care facilities, 10 video demonstrations of the robot's behavior were presented to the physicians at Sant'Andrea Hospital.

The recipients we asked to rate how satisfied they were with the robot's behavior in each demonstrated scenario on a 5-point grading scale, with 1 representing absolutely unacceptable behavior and 5 representing perfection. The feedback form was filled out by 40 participants, and their average acceptance results are reported in aggregated form in \autoref{table:acceptancegen}.

The  \autoref{table:acceptancegen} shows that the average level of robot acceptability is high and can be rounded up to $82\%$, regardless of the nature or difficulty of the simulated scenario. Thus, domain experts were not predisposed to higher cautiousness within more domain-specific or chaotic settings, resulting in a positive outlook on the robot's future integration in a safety-critical environment. Furthermore, average robot acceptability and the standard deviation from the
average computed for each estimated scenario separately confirm the hypothesis of clustered, scenario-independent results, as none of the average values differs significantly from the others.

\begin{table}[!htb]
\centering
\caption{\centering \textbf{Average robot acceptability:} Avg $-$ in entire population, Avg Med $-$ in medical contexts, Avg Non-Med $-$ in common sense contexts, Avg Crowded $-$ in crowded environments, and Avg Non-Crowd $-$ in environments with 1 dynamic human.}
\label{table:acceptancegen}
\begin{tabular}{| l | r |}
 \hline  
  \textbf{Metric} & \textbf{Value[\%]}\\ [0.1ex]
 \hline
 Avg & 81.7\\\hline
 Avg Med & 81.5\\\hline
 Avg Non-Med & 81.9\\\hline
 Avg Crowd & 81.5\\\hline
 Avg Non-Crowd & 81.8\\\hline
\end{tabular}
\end{table}

\begin{figure*}[!h]
    \centering
    \includegraphics[width=.9\linewidth]{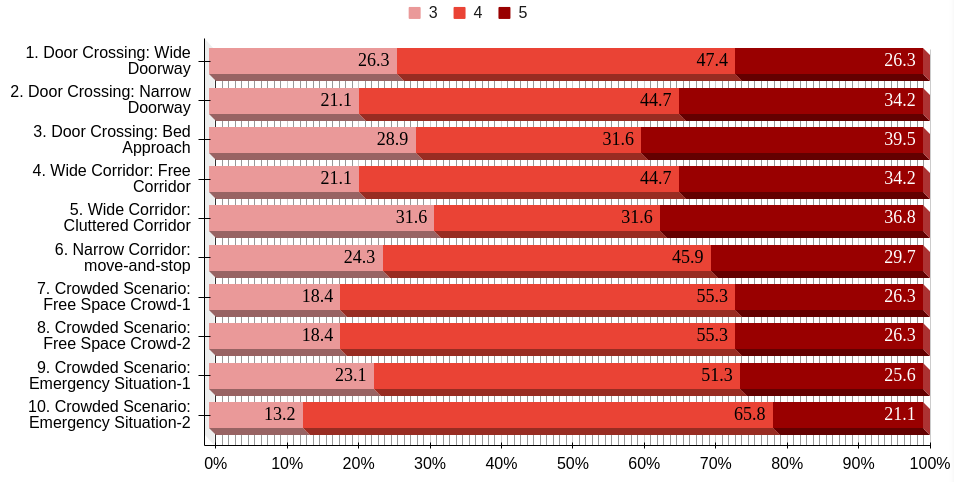}
    \caption{\centering The percentage distribution of grades in each scenario presented to health-care workers.}
    \label{fig:grades}
\end{figure*}

It is worth noting that the lowest encountered estimate in all demonstrated scenarios was 3, indicating that nobody of the participating providers rated the robot's behavior as more likely to be unacceptable than acceptable. And another indicator of consistency is the median value, which is 4 in all cases and is equal to the mode value in all cases, except for the Door Crossing: Bed Approach and Wide Corridor: Cluttered Corridor where the mode is 5. To gain more intuition on the distribution of the grades assigned by experts to each video simulation let us display the bar chart in \autoref{fig:grades}.

A precise examination of the chart leads to the conclusion that the robot's behavior cannot be considered more acceptable in some cases than in others. For example, the two previously mentioned scenarios: Door Crossing: Bed Approach and Wide Corridor: Cluttered Corridor, have the highest concentration of both minimum and maximum given estimates, 3 and 5, which does not allow to mark these scenarios as neither preferred nor disregarded by clinicians.

Overall, it seems from the chart in \autoref{fig:grades} and from the calculated statistics in general that even confusing and causing inconveniences for humans robot behaviors such as following behind the person (Crowded Scenario: Free Space Crowd-1), stopping at random location in the middle of the crowd (Crowded Scenario: Emergency Situation-2) or blocking the human's way (Narrow Corridor: move-and-stop human) would be tolerated by people. As a result, expert validation of the robot's performance leads to a positive conclusion about the possibility of human-robot coexistence in health-care facilities.

\section{Conclusion and Future Work}
\label{sec:6}
Human-Aware Navigation is a field that broadens the horizons of robotics research by empowering existing robotic systems with socially meaningful capabilities and contributing to the seamless integration and sustainability of robots in human environments. It proposes a wide range of challenging tasks associated with the incorporation of the socially enhanced components into established navigation frameworks. One of such tasks is the enhancement of robot reliability in the healthcare sector, which we address in this work. In our case, the socially enhanced component is a Co-operative Human-Aware Navigation (CoHAN) planning system, and the established navigation framework is MARRtina robot software.

The results of the comparison of the integrated system with a non-social planning approach manifest an improvement in robot reliability, and the qualitative analysis demonstrates that the robot behavior became more socially compliant, which, in turn, promotes human acceptance of the robot, as confirmed by the statistical analysis of human validation of the results.

Furthermore, the work done on integration into MARRtina of a human-aware planning system opens the scope of future investigations for the community of people who are interested in using this robot for their own studies and developments in the field of Human-Aware Navigation.

Then, probably, the greatest value of this work lies in a thorough examination of the planner's features. While acknowledging the limitations of the system's potential applications, it also provides intuition on the directions of future system-upgrading initiatives.

A room for improvement has been found in the resilience of the \verb|Backoff-recovery| method and further development of its prototypical edition. Then, the goal selection in the \verb|PredictGoal| service can be automated by the implementation of some probabilistic goal inference approach. Besides, new Human Path Prediction techniques can be embedded into system to meet a need for improved estimation of human motion that has been verified throughout experimentation. The experimentation also suggests that the human model has to be updated. Eventually, the proactive and anticipative behavior on the side of the robot are two particularly important aspects, and they are even more critical in medical settings. Therefore, the inclusion of new methods with a design focus on strengthening the system in these aspects would be beneficial.

Shading more light on the vectors of the system upgrades, we plan a re-integration of the current version of the planner with a newer one \cite{ref:multi}. The key motivation for doing so is to boost the robot's capability to act proactively by accounting for humans outside the Visible Region.

In addition, the system improvement challenge can be approached from a completely different angle: the human controller module can be enhanced, for instance, by creating human avatars that demonstrate rational social behaviors rather than just moving in a primitive way according to predefined rules. The insightful works on that are presented by Favier et al. in \cite{ref:rel28} and \cite{ref:rel29}. The authors propose a system called InHuS, which incorporates autonomous, intelligent human agents specifically designed to act and interact with a robot navigating in a simulated environment. Since this system is based on CoHAN, it would be interesting to combine it with the already integrated planner in the MARRtina software.

Finally, in order to complete a characterization of the system, we intend to deploy it on a physical robotic platform within a real-world clinical environment. As a result, new challenges will likely be identified, necessitating the development of new solutions.

\bibliography{sample-ceur}

\appendix





\end{document}